\documentclass[runningheads]{llncs}
\usepackage{graphicx}
\usepackage{comment}
\usepackage{amsmath,amssymb} 
\usepackage{color}
\usepackage{epsfig}
\usepackage{graphicx}
\usepackage{mathrsfs}
\usepackage{hhline}
\usepackage{subfig}
\usepackage{mathptmx}
\usepackage{booktabs, makecell}
\usepackage{hyperref}
\hypersetup{colorlinks= true,
            linkcolor=red}

\begin{document}
\pagestyle{headings}
\mainmatter
\newcommand{\red}[1]{\textcolor{red}{#1}}

\title{Graph2Kernel Grid-LSTM: A Multi-Cued Model for Pedestrian Trajectory Prediction by Learning Adaptive Neighborhoods} 


\titlerunning{Graph2Kernel Grid-LSTM}
\authorrunning{S. Haddad and S.K. Lam}
%
\author{Sirin Haddad\inst{1}
\and
Siew-Kei Lam\inst{2}}
%
%
\institute{Nanyang Technological University, 50 Nanyang Ave, 639798, Singapore\\
\email{siri0005@e.ntu.edu.sg}\inst{1} \quad
\email{assklam@ntu.edu.sg}\inst{2}}



\def\httilde{\mbox{\tt\raisebox{-.5ex}{\symbol{126}}}}

\maketitle

\begin{abstract}
    Pedestrian trajectory prediction is a prominent research track that has advanced towards modelling of crowd social and contextual interactions, with extensive usage of Long Short-Term Memory (LSTM) for temporal representation of walking trajectories.
    Existing approaches use virtual neighborhoods as a fixed grid for pooling social states of pedestrians with tuning process that controls how social interactions are being captured. This entails performance customization to specific scenes but lowers the generalization capability of the approaches. In our work, we deploy \textit{Grid-LSTM}, a recent extension of LSTM, which operates over multidimensional feature inputs. We present a new perspective to interaction modeling by proposing that pedestrian neighborhoods can become adaptive in design. We use \textit{Grid-LSTM} as an encoder to learn about potential future neighborhoods and their influence on pedestrian motion given the visual and the spatial boundaries. Our model outperforms state-of-the-art approaches that collate resembling features over several publicly-tested surveillance videos. The experiment results clearly illustrate the generalization of our approach across datasets that varies in scene features and crowd dynamics.
    \footnote{\href{}{Code: https://github.com/serenetech90/multimodaltraj\_2}{}}
\end{abstract}

\section{Introduction}
Predicting pedestrian trajectory is an essential task for accomplishing mobility-based jobs in naturalistic environments, such as robotics navigation in crowded areas, safe autonomous driving and many other applications that require foreseeing motion in an interactive dynamical system. Existing literature has focused so far on modeling the social interactions between pedestrians by discretizing the environment as a grid of local spatial neighborhoods \cite{alahi2016social}, taking global scope of the whole scene \cite{vemula2018social} or inferring relationships between pedestrians pairwise \cite{choi2019looking}. So far, spatial neighborhoods have been the fundamental basis for considering pedestrians influence on each other, accounting only for the positional and higher-order motion features. 
Other methods have used additional features such as head pose \cite{hasan2018mx,hasan2019forecasting,hasan2018seeing} to attain the visual field of attention in pedestrians in order to assess how they are related to each other.  

Classical approaches \cite{hasan2018seeing,helbing1995social,yamaguchi2011you} resorted to a hardcoded quantitative threshold to define the proximal distance for neighborhoods. The use of deep learning improves the neighborhood concept by learning a custom threshold for each pedestrian \cite{kipf2018nri}. 
MX-LSTM \cite{hasan2018mx} determines neighborhoods around pedestrians visually based on the fixed visual horizon of pedestrian. Within this imaginary area, pedestrians spatial states get pooled. To this end, MX-LSTM only considers the correlation between head pose and speed. In the aforementioned works, generalization can be problematic as the models rely on fixed assumptions that pertain to specific scenes. Recent works introduced additional features such as the looking angle of pedestrians \cite{hasan2018mx}. These approaches and GAN-based approaches resorted to dedicating several pooling layers for treating multiple features \cite{lisotto2019social,ridel2019scene} or multiple trajectories states \cite{amirian2019social,sadeghian2018sophie,zhang2019seabig} which will require additional transformation and pooling layers or assigning separate LSTM to each pedestrian in order to treat all the features together.

\begin{figure}
    \centering
    \includegraphics[width=6cm,
    height=4cm, trim={0 0 0 2cm }]{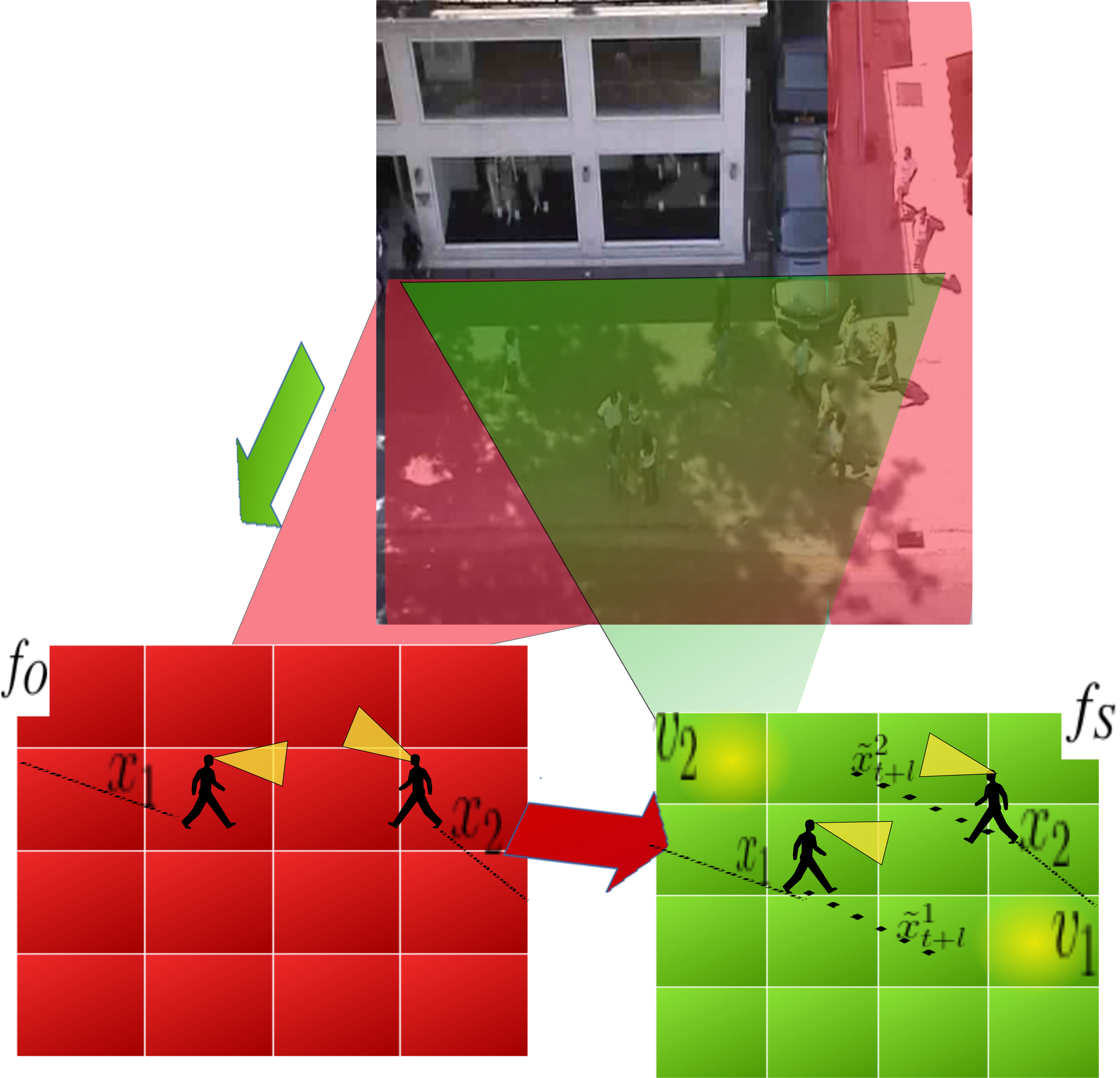}
    \caption{Illustration of adaptive deep neighborhoods selection process. The process is comprised of two grids. To the left, the static neighborhood grid $f_O$ segments the scene image into several local regions. It takes pedestrians looking angle $v_1, v_2$ (shaped as yellow cones) to stem their awareness of the static surroundings. The dynamic grid $f_S$ takes pedestrians trajectories $x_1, x_2$ along with their looking angle to stem their social interactions. The end goal is to predict future trajectories $\Tilde{x}^1_{t+l}, \Tilde{x}^2_{t+l}$ accurately given the estimation of future potential neighborhoods and the best modeling of social interactions. To the right, the output static grid has few highlighted areas, which indicates future neighborhoods where pedestrians would walk. Also note the leaning links connecting pedestrians to indicate how the existence of social influence on each other.} 
    \label{fig:intro_fig_visuospatial_ngh}
\end{figure}{}


In order to overcome the limitations of existing approaches, we estimate the neighborhood given the static context around pedestrian and a mixture of social cues that defines pedestrian situational awareness and social interactions. More concretely, we define an adaptive neighborhood that relies on visual locus and spatial constraints, supported by the existential correlation that associates the head pose to walking direction and intention. This is based on the observation that although people often do not articulately plan their walking trajectories, they keep their attention focused within coarse path boundaries. They also do not strictly define their neighborhoods. The neighborhood is a virtual concept to discretize the scene and depends on a mixture of social cues that indicate social interaction state and pedestrian awareness.
Our approach updates the neighborhood definition for a multi-cued context, where multiple features are encoded for pedestrians. We coin this Visuospatial neighborhoods, as such neighborhoods characterize the spatial distance around pedestrians as well as their visual attention boundaries. 
Figure \ref{fig:intro_fig_visuospatial_ngh} shows that neighbourhood $\nu_4$ is highlighted with yellowish shade. This is perceived as a future neighborhood for pedestrians that are related to each other. Hence their states are to be pooled together as they share the same future neighborhood. Anticipating some region as potential neighborhood relies on where they look and their walking path.

We propose Graph-to-Kernel LSTM (G2K LSTM) that combines the spatial physical boundaries and the visual attention scope to shape each pedestrian neighborhood. G2K LSTM is an LSTM-based approach that transforms a spatio-temporal graph into the kernel to estimate the correlations between pedestrians. This correlation represents the importance of a relationship between two people, stemming from the natural correlation between their visual angle and their relative distances. In sociological and psychological proxemics \cite{hall1966hidden,bera2017sociosense}, human maintains specific distance from others to be his/her personal zone within which they feel comfortable and evade collisions. This zone is maintained by what they observe and pay attention to.

Attaining accurate estimation of such neighborhoods is a non-trivial task, due to the stochasticity pertaining to the ground-truth definition. Following a stochastic optimization, neighborhood formation is bounded to minimize the trajectory prediction errors, such that the underlying graph structure improves the social interaction modeling. 
We develop neighborhood modeling mechanism based on Grid-LSTM \cite{kalchbrenner2015grid}, which is a grid mask that segments the environment into a regular-shaped grid. It encompasses the sharing mechanism between adjacent neighborhoods in the grid. This sharing determines how pedestrians social interaction is modeled in deep networks. Predicting pedestrians future trajectory can be used to further define future neighborhoods, thereby leading to a better understanding of how pedestrians are related and influenced by each other.

In summary, this paper delivers the following contributions:
\begin{itemize}
\item We introduce Grid-LSTM into pedestrian trajectory prediction to encode multiple social cues altogether. Seeding from the natural correlation found between the visual and the spatial cues mixture, the network learns a soft-attention mechanism for evaluating given social interaction importance from within the provided data. The correlation between head pose and walking direction emphasizes pedestrian walking intention and in this work we combine head pose with walking trajectories to improve social and contextual modeling of pedestrians interactions.

\item We present a deep neighborhood selection method for estimating the influence of social relationships between pedestrians. It guides the message-passing process according to a relational inference in the graph-based network that decides the routes for sharing the attention weights between pedestrian nodes. 

\item Due to the aforementioned data-driven mechanisms, our approach yields state-of-the-art results on widely-tested datasets. Our models also produce consistent results across the datasets, which indicates its generalization capability to various crowd contexts. 
\end{itemize}



\section{Related Work}
\paragraph{The visual field and the situational awareness.}

Existing works have shown the benefits of combining head pose with positional trajectory for prediction. The head pose is used as a substitute for the gaze direction to determine the Visual Field of Attention (VFOA) \cite{hasan2018seeing,yang2018my}.
The head pose feature correlates with the walking direction and speed, which 
emits pedestrian destinations as well as their awareness of the surrounding context. The visual field of attention in pedestrians relied on assumptions that align head pose with gaze direction to fixate the attention region as pedestrian is walking. In resemblance to \cite{yang2018my}, we argue that the width of the visual field and its shape shall affect the representation of pedestrian visual awareness state and thereby their neighborhood perception.

To a large extent, when pedestrians are walking, they only consider other pedestrians who are close and can pose a direct influence. This social situation can be captured when using a narrow visual angle, i.e. 30 $\deg$. Nevertheless, pedestrian does not always look straight, they tend to tilt their heads and therefore perceive more about the environment structure and other dynamical objects around them. Using a wider sight span for each pedestrian allows a better perception of pedestrian awareness and focal attention.



\paragraph{Relational inference for neighborhood selection.}
Extensive research is conducted on relationships inference between entities of data in image segmentation \cite{scarselli2008graph,battaglia2018relational}, graph recovery for interactive physical systems \cite{webb2019factorised,kipf2018nri}. Recently, researchers engaged spatio-temporal graphs to model pedestrians relationships to each other \cite{jain2016structural,vemula2018social}. More advanced approaches such as \cite{choi2019looking,sadeghian2018sophie,xue2019location,fernando2018soft+} evaluate the relational importance between pedestrians using neural attention technique. In addition to using attention, \cite{zhang2019sr} deploys the refinery process that iterates over variants of the neighborhood until it selects the best neighbors set for each pedestrian. In our work, we similarly target neighborhood selection problems but with fewer iterations and faster recurrent cell called Grid-LSTM that requires fewer training epochs. 





\paragraph{Social neighborhoods for human-human interaction.}
In the literature, this area is extensively studied as two separate disciplines: Human-human interaction and Human-space interaction. Forming neighborhoods is found to be based on a single type or a combination of both interaction types and the objective is to determine a way for combining pedestrians and discovering their influence. The approaches that socially define the basis for neighborhoods \cite{alahi2016social,cheng2018pedestrian,xue2018ss} proposed pooling techniques to effectively combine spatially proximate pedestrians, while \cite{haddad2020self} proposed influence-ruled techniques to combine pedestrians that align their motion according to each other. However, even when modeling the spatial neighborhoods with the aid of context information, existing approaches generally involve hardcoded proxemic distance for outlining neighborhood boundaries as fixed grids. While these works achieve successful results, they fail to consider dynamic environments representation which may cause failure in cases that require adaptive neighborhoods or extra cues such as pedestrian visual sight span, to ascertain the conjunction between social motion and contextual restrictions. 

\section{Our Approach}

\subsection{Problem Formulation}

The problem of learning neighborhood boundaries is formulated as minimization of Euclidean errors between the predicted trajectory $\widetilde{X}$ and ground-truth trajectory $X$:
\begin{equation} \label{eqn_costfn}
     \mathcal{L} = \underset{J_\theta}{argmin} ||\quad \widetilde{X} - X \quad||^2_2
\end{equation}{}

Such that $J_\theta$ refers to the network trainable parameters that minimize the loss, $\mathcal{L}$, \textit{L2-norm} of $\widetilde{X}$ and $X$.

Let $X$ be pedestrians trajectories, such that: $X$ = ${x_1, x_2, ... , x_n}$, with n pedestrians. $\widetilde{X}$ are future trajectories, $x^{i}_t$ is i-th pedestrian trajectory from time-step $t = 1$ until $t = t + obs$, given that $obs$ is observation length. We observe 8 steps of each pedestrian trajectory and predict for the next 12 steps.
Each predicted step is added to the first predicted point, $\Tilde{x}^i_{t+pred} = \Tilde{x}^i_{t+1} + \Tilde{x}^i_{t+2} + ... + \Tilde{x}^i_{t+l}$ to maintain consistency and dependency between predicted steps.


\subsection{Temporal Graphs}
Given a set of pedestrians P, we represent their trajectories over time using a temporal graph $G_t$ at each time step $t$, containing \ensuremath{\mathcal{N}} nodes, such that each pedestrian is assigned a node $n_t$ to store their position and temporal edge $e_t$ that links the node $n_{t-1}$ to $n_t$. 

Our approach Graph-to-Kernel (G2K), maps temporal graphs into a kernel of fixed dimension, $K$, such that the kernel generates predicted trajectories $\widetilde{X}$ and adjacency states between the pedestrian nodes. Adjacency models the social interactions. So $J_\theta$ in Eq. \ref{eqn_costfn} refers to pedestrians associations to each others:
\begin{equation}\label{eqn:kernel}
    \widetilde{X} , J_\theta = K(G_{(\mathcal{N},\nu)})
\end{equation}{}


\subsection{Social Relational Inference (SRI)}
\label{sec:sri}
In this section, we explain the SRI unit which is the proposed kernel. We detail the steps for the mechanism in which the kernel $K$ performs deep relational inference between pedestrians. SRI unit starts estimating the adjacency state for each pedestrian. It forms the adaptively-shaped social neighborhoods.

The kernel has a customized design relevant to the features set included in our model versions.
As defined earlier in Eq. \ref{eqn:kernel}, kernel $K$ generates the social interaction states and future positional predictions, so according to $MC$ model and $MCR^*$ models set, Eq. \ref{eqn:kernel} is updated as follows:
\begin{equation}
   \label{mc_mcr_K}
    \widetilde{X}, J_\theta = K(f_S, V, h)
\end{equation}{}

A single Grid-LSTM cell $NLSTM_{\nu}$ is used as the encoder of the social human-human interactions. It assumes that motion happens over the uniformly-divided square grid where each neighborhood is tagged by $\nu$.

SRI casts $NLSTM_{\nu}$ over the temporal graph $g_t$. It takes pedestrians trajectories, $X$, initial hidden state $h$ and generate embeddings of spatial features $f_S$. To formulate multi-cued trajectories set $T$, we firstly encode positional trajectories $X$ using two-stages transformation function $\phi$ as follows:
\begin{equation}
    \mathcal{X} = \phi (X)
\end{equation}
\begin{equation}
    \phi (X) = W_{ii} * (W_i * X)
\end{equation}

In our work, we coin a new term for multi-cued neighborhoods as "Visuospatial" neighborhood, that is a combination of pedestrians spatial whereabouts and their visual sight span.

To represent the Visuospatial neighborhoods, we encode the 2D head pose annotations, also called Vislets, $V$ using single-stage transformation function $\phi$ as follows: 
\begin{equation}
    \mathcal{V} = \phi (V)
\end{equation}
\begin{equation}
    \phi (V) = W_i * V 
\end{equation}

Then we concatenate the embedded cues $\mathcal{V}$ and $\mathcal{X}$ in Eq. to be fed into the GNN Grid-LSTM in Eq. \ref{eqn:ngh_glstm}:

\begin{equation}
    T = (\mathcal{X}, \mathcal{V})
\end{equation}{}
\begin{equation}
    \label{eqn:ngh_glstm}
    f_S, h_S = NLSTM_{\nu}(T, h_S)
\end{equation}{}

We developed several models starting with the simplest model comprising of single Grid-LSTM cell with only positional trajectories. We term it as G-LSTM model. 
According to G-LSTM model, kernel $K$ of Eq. \ref{eqn:kernel} is updated as follows: 
\begin{equation}
    \label{glstm_K}
    \widetilde{X}, J_\theta = K(f_S, h)
\end{equation}

\begin{figure}
    \centering
    \includegraphics[width=0.85\linewidth, height=6cm]{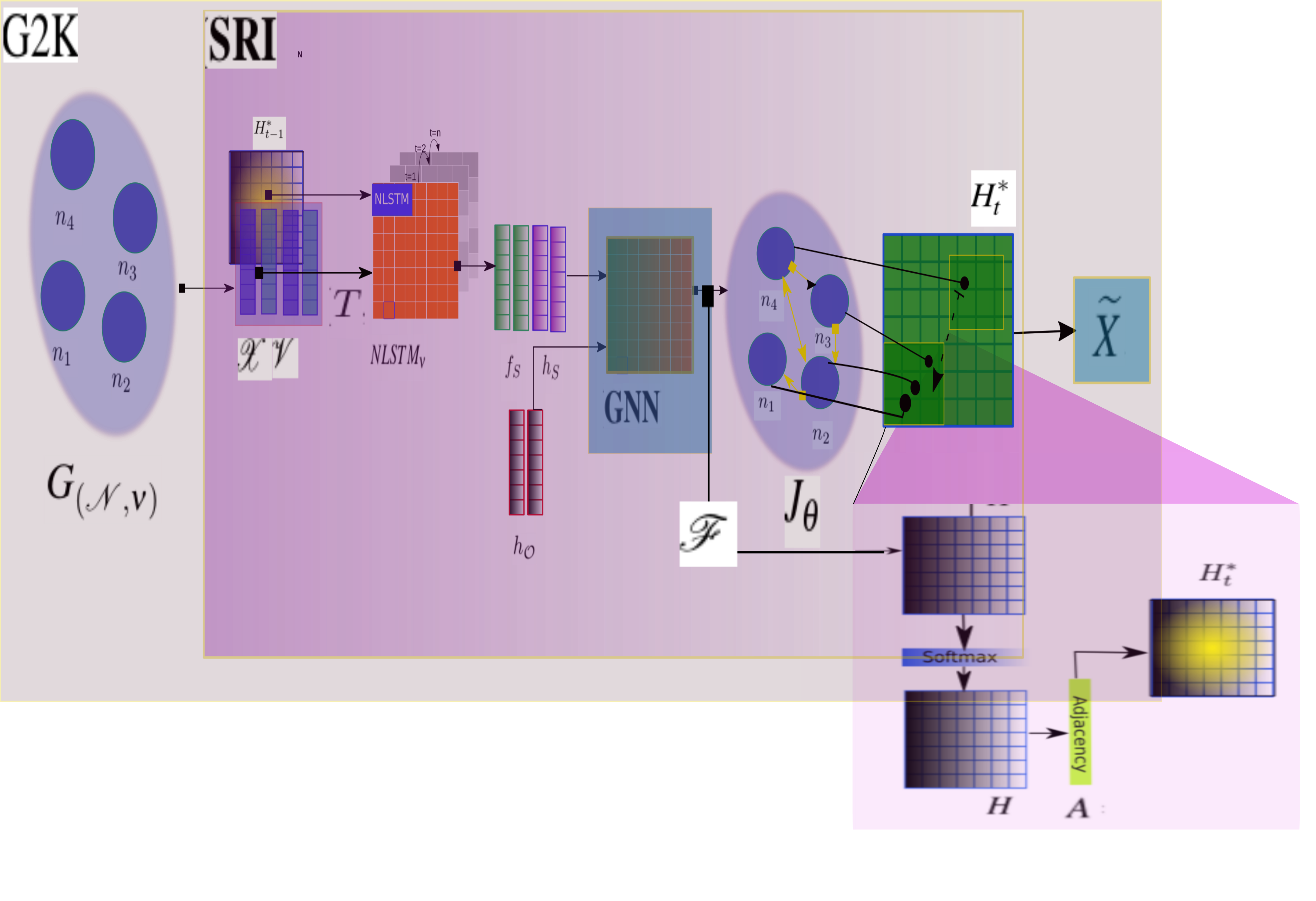}
    \caption{Full pipeline of G2K kernel. The SRI network encodes Vislets $\mathcal{V}$ and positional trajectories $\mathcal{X}$ for each pedestrian trajectory. Then maps them into social grid mask $f_S$ using $NLSTM_\nu$. The GNN network discretize static context using $NLSTM_O$ into 'Visuospatial' neighborhoods and stores pedestrian contextual awareness in $f_O$. At the consequent step, SRI takes $f_O$ and $f_S$, and maps them into the weighted adjacency matrix. This will generate the edge set $\nu$ as means of completing graph $G_t$ at time-step $t$.}
    \label{fig:sri}
\end{figure}{}

After that, we develop two versions of Multi-Cued Relational inference (MCR) model: $MCR_n$ \& $MCR_{mp}$. 
Hidden states sharing between pedestrians is guided by a deep relational mechanism that involves learning the social interaction effect. This effect is manifested through soft-attention that evaluates pedestrian relationships importance and chooses the more influential relationships set than other associations that can have a weaker impact on pedestrian trajectory. Consequently, the hidden states become messages passed between pedestrians based on a deep selective process that determines who are the neighbors of each pedestrian as displayed in Figure \ref{fig:sri}.


As part of the SRI network, we run GNN network to produce the final static features $\mathcal{F}$. SRI takes this output and feeds it through another transformation function $\phi$. The latter embeds the features set output by GNN unit and yields $\mathcal{F}'$:
\begin{equation}
    \mathcal{F}' = \phi(\mathcal{F})  = C * ( (W_v * [f_S, \mathcal{V}] + b_v) * (W_r * \mathcal{F}) )
\end{equation}
In $MCR_{mpc}$, $\mathcal{F}$ stores the enhanced modeling, comprised of physical spatial constraints $C$ and social relative features $f_S$.
In $MCR_n$, $\mathcal{F}$ will only take the social relative features $f_S$.

SRI unit will tune the influence of each region. It deploys scaled self-attention mechanism \cite{velivckovic2017graph} (as defined in Eq. \ref{eqn:soft_attn}). Attention coefficients $\mathscr{a}$ considers the human-human interaction features and the human-space interaction features:
\begin{equation} \label{eqn:soft_attn}
    a =  \frac{Softmax (\exp{(\mathcal{F}')})}{\Sigma \exp{(\mathcal{F}')}}
\end{equation}{}


Finally, $MCR_n$ kernel neurally evaluate hidden states by passing through Softmax layer. This will transform pedestrians nodes states into the continuous space of the interval [0,1]:
\begin{equation} \label{eqn:softmax_fnri}
     Softmax(H) := [0,1]
\end{equation}{}

Given the $i$th pedestrian trajectory features, neighborhood $\nu^i_t$ of pedestrian $i$ is defined as follows:
\begin{equation}
    \nu^i_t =  \{(n_i, n_j)\}^+; \quad |\nu^i_t| <= |\mathcal{N}|
\end{equation}{}

Moreover, neighborhood boundaries at pedestrian $i$ are defined by the set of edge pairs that connect pedestrian nodes to other nodes in the graph.
Note here that $\nu$ is the Greek symbol (\textit{Nu}) and it is different from the symbol $V$ which was earlier tagged to the Vislets.

Inspired by the neural factorization technique \cite{webb2019factorised} to factorize spatio-temporal graph edges, we establish a deep mechanism for neighborhoods that is aware of the static and the social constraints.
Since the best setting for neighborhoods is unknown, we estimate pedestrians relationships using their relative visual attention and spatial motion features to learn their neighborhoods. 

In $MCR_{mp}$ kernel, the model directly factorizes relationships between pedestrians by passing importance weights in Eq. \ref{eqn:soft_attn} as a message with the hidden states matrix and thresholding the strong and weak relationships:
\begin{equation}
    H = Softmax (f'_{\mathcal{O}_{t+1}}*H)
\end{equation}{}

For accomplishing social relational inference of pedestrians interactions, the last two steps are common among models: $MCR_n$, $MCR_{mp}$ and $MCR_{mpc}$. Using a normally-initialized linear transformation matrix $W$, the new hidden states are mapped into adjacency matrix to determine the edges in the graph which manifest pedestrians relationships with various degrees of strength:
\begin{equation}
    A = W*H
\end{equation}{}

Eventually, $J_\theta$, is the updated pedestrian adjacency states $H^*_t$, which will be fed again in the following time-step to the social neighborhood encoder in Eq. \ref{eqn:ngh_glstm}: 
\begin{equation}
    H^*_{t} = A*H
\end{equation}{}


The resulting adjacency-based representation $H^*_{t}$ is evaluated by whether the new proposal of edge set, generates better modeling than the previous time-step edge set. 
This is set as the objective cost function $J_\theta$. 

\subsection{Gated Neighborhood Network (GNN)}
\label{sec:gnn}
In this section, we explain our proposed mechanism for deep neighborhoods formation.

GNN unit involves a Grid-LSTM \cite{kalchbrenner2015grid} cell $NLSTM_O$ to encode contextual interactions. It assumes the scene as fixed-shaped grid of shape $[n x n]$. 

\begin{figure*}
    \centering
    \includegraphics[width=\linewidth, height=6cm]{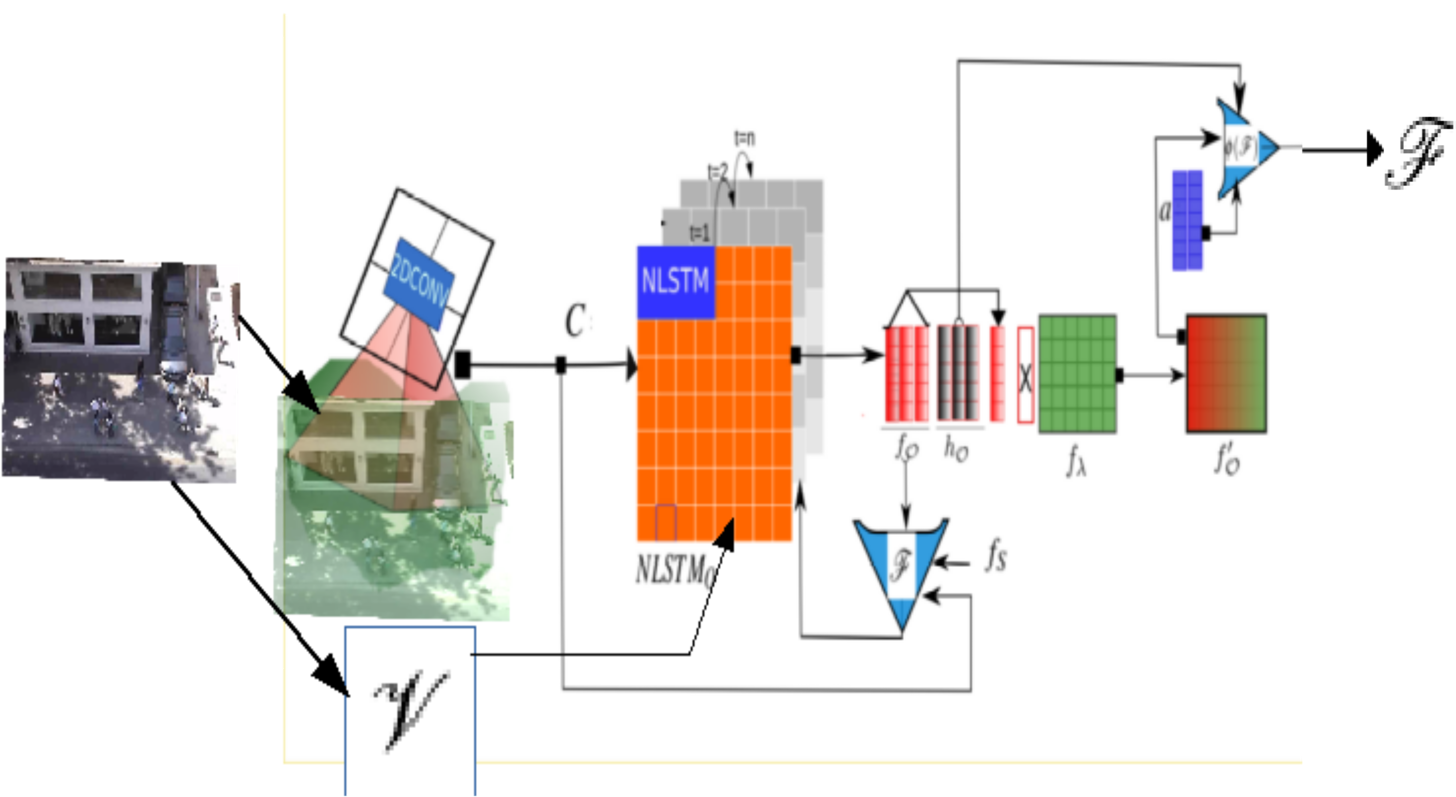}
    \caption{Gated Neighborhood Network pipeline. At the beginning, $2DCONV$ encodes a static image of the scene and forward the features into NLSTM cell which discretizes the environment into a virtual grid.}
    \label{fig:gnn}
\end{figure*}{}


Figure \ref{fig:gnn} illustrates the pipeline of GNN. $2DCONV$ is a 2D convolutional layer used for encoding the contextual interaction. It runs a grid mask filter $M$ and assumes that static space neighborhoods are discretized into a square grid. The mask filter is a normally-initialized matrix to indicate that initially, all local regions are of equal influence on pedestrian motion. 
\begin{equation}
  C = 2DCONV(f_S, h_\mathcal{O}, M)
\end{equation}{}

Compared to literature approaches, the Social-Grid model \cite{cheng2018pedestrian} dedicates a separate Grid-LSTM  \cite{kalchbrenner2015grid} for each pedestrian. In our models, we use only a single Grid-LSTM to encode a set of features for all pedestrians. 

MX-LSTM \cite{hasan2018mx} applies the concept of Visual Field of Attention (VFOA) to determine pedestrians attention to each other. They hard-code pedestrians VFOA and rely only on the head pose which to some extent can validate the looking angle. 

The social pooling and the visual pooling proposed in \cite{alahi2016social,hasan2018mx} respectively selects one feature modality for learning about pedestrians interactions by pooling their states into pre-determined neighborhoods. 

In our work we encode head pose $V$ with static features $C$ to formulate the "Visuospatial" neighborhood representation as means of pedestrian attention to static context, using single Grid-LSTM cell $NLSTM_O$:

\begin{equation} \label{eqn:visuospatial}
     f_\mathcal{O}, h_\mathcal{O} = NLSTM_O(C, V, h_\mathcal{O})
\end{equation}{}

Eq. \ref{eqn:visuospatial} introduce multi-modality concept through combining both, visual awareness state $V$ and physical spatial constraints $C$, for stemming pedestrians attention to the physical context. In other words, encoding pedestrians interaction with the static context is stored in $f_\mathcal{O}$. Later in Section \ref{sec:sri}, $f_\mathcal{O}$ will be passed with the social interaction features $f_{S}$ to calculate Adjacency state of each pedestrian.

Taking the static grid features $f_{\mathcal{O}}$ in Eq. \ref{eqn:visuospatial} are regularized by a constant factor of $\lambda$ as a means of starting with equally important static neighborhood regions:
\begin{equation}
    f_{\mathcal{O}}^\ensuremath{\prime} = f_{\mathcal{O}} * f_{\lambda}
\end{equation}{}


In formal definition, given a frame $f$, we discretize it into a grid of uniformly-shaped grid $G$ of $k$ neighbourhoods. Initially, coarse neighborhoods in the grid $f_\mathcal{O}$ are assigned equal importance which will be adapted through training batches.

A final neighborhood mask $\mathcal{F}$ contains a combination of fixed and relative pedestrians trajectory features, $f_O \& f_S$. $\mathcal{F}$ can vary depending on the input features to the static mask:
\begin{equation}
    \mathcal{F} = f_S * f_{\mathcal{O}}^\ensuremath{\prime}
\end{equation}{}


The static features $f_\mathcal{O}$ are weighted by soft-attention coefficient $a$ using the scaled soft-attention mechanism \cite{bahdanau2014neural}, before being neurally evaluated in Eq. \ref{eqn:soft_attn}:
\begin{equation}
    f'_{\mathcal{O}_{t+1}} = a * f'_{\mathcal{O}_{t+1}} * h_\mathcal{O}
\end{equation}{}



    


\section{Experiments} \label{sec:exp}

\subsection{Accuracy Metrics}

We use Euclidean average errors same as in \cite{hasan2018mx,alahi2016social}, which are:
\begin{itemize}
    \item Average Displacement Error (ADE):
        measures prediction errors along the time-steps between the predicted trajectory and the ground-truth trajectory as follows:
    \begin{equation}
          \frac{\sum_{i = 1}^N \sum_{j = 1}^l ||(\widetilde{X_i^j} - X_i^j)||_2 } {N * |T|} ,
    \end{equation}{}
    \item Final Displacement Error (FDE): measures prediction errors at the final time-step between the predicted trajectory and the ground-truth trajectory as follows:
    \begin{equation}
            \frac{  \sum_{i = 1}^N ||(\widetilde{X}^{T-1}_i - X^{T-1}_i)||_2 } {N} \quad .
    \end{equation}{}
\end{itemize}

\subsection{Implementation Details}

During training, we set the batch size to 16, grid size to 4 and lambda parameter to 0.0005. We set the hidden size to 128 for Grid-LSTM, the number of frequency blocks to 4, frequency skip to 4 and the number of cell units to 2. Each frame gets segmented into 8 virtual zones as indicated by the grid size and set the global neighborhood size to 32. In all experiments over our models, the features extracted from the static grid and social neighborhood vector were mapped to hidden states of length fixed at 10. With Grid-LSTM we set the training epochs to 10 as we noticed that LSTM has reached convergence within that number of iterations without any major improvements in the learning curve beyond that.

\subsection{Baseline Models and Proposed Models}

\begin{itemize}
    \item \textit{S-LSTM} \cite{alahi2016social}. Dedicates LSTM for every pedestrian, and pool their states before predicting future steps. Their method only combines features of pedestrians who are found occupants of common neighborhood space. The neighborhood and occupancy grid sizes are set empirically for attaining the best results over ETH and UCY datasets.
    \item \textit{MX-LSTM}. A multi-cued model that encodes Vislets and Tracklets to estimate pedestrian trajectory using LSTM. It determines pedestrians relationships using fixed visual frustum that indicates their visual attention state.
    
    \item \textit{SR-LSTM} \cite{zhang2019sr}. A LSTM-based network inspired by message-passing graph neural networks \cite{gilmer2017neural} which improves pedestrians motion representation and social neighborhoods for predicting future trajectory over fully-connected spatio-temporal graphs. 
    
    \item \textit{G-LSTM}. Single Grid-LSTM cell that encodes pedestrian walking trajectories without additional cues. It doesn't include Sections \ref{sec:gnn} and \ref{sec:sri} in its pipeline.
    
    \item $MC$ Multi-Cue inference.
    This model combines Vislets and Tracklets to predict pedestrians trajectories. It uses Vislets to incorporate pedestrian situational awareness into their motion modeling, however, it does not differentiate pedestrians influence. 
    
    \item $MCR_n$ Multi-Cue Relational inference.
    Built upon the $MC$ model, it capitalizes on its understanding of situational awareness to differentiate pedestrians importance on each other for more accurate modeling of social interactions. 
    
    \item $MCR_{mp}$ Multi-Cue Relational inference. Uses message passing mechanism through virtual static neighborhood grid that directly passes importance values which reflects pedestrians social interaction.
    
    \item $MCR_{mpc}$. In addition to the previous design components, this model encodes the visual static feature of the scene using a 2D convolutional layer to account for the contextual interaction between pedestrian and scene structure.
    
\end{itemize}{}
\subsection{Quantitative Analysis}





\begin{table*}[t]
    \begin{center}
    \begin{tabular}{ccccccc}
    \hline
     Metric & Model & Zara1 & Zara2 & UCY-Univ& TownCentre & AVG\\ \hline
     ADE & S-LSTM \cite{alahi2016social} & 0.68 & 0.63 & 0.62& 1.96 & 0.64 \\
     & MX-LSTM \cite{hasan2018mx} & 0.59 & 0.35 & 0.49 & 1.15 & 0.48 \\ 
     & SR-LSTM\cite{zhang2019sr} &  \textbf{0.41} & \textbf{0.32}  & 0.51 & 1.36 & 0.65 \\
     & MC & 0.46 & 0.38 & 0.47 & 0.43& 0.44 \\
     & MCR\_n & 0.47 & 0.38 & 0.47 & 0.39 & 0.43\\
     & MCR\_{mp} & 0.45 & 0.38 & \textbf{0.45} & \textbf{0.34}& \textbf{0.41}\\
     & MCR\_{mpc} & 0.45 & 0.38 & 0.47 & 0.39 & 0.42\\\hline
    Metric& Model & Zara01 & Zara02& UCY-Univ &TownCentre & Avg \\\hline
    FDE & S-LSTM \cite{alahi2016social} &1.53 & 1.43&	1.40 &3.96 &1.45 \\
     &MX-LSTM \cite{hasan2018mx}&1.31 & 0.79 & 1.12 & 2.30 &	1.38 \\
    & SR-LSTM\cite{zhang2019sr}&\textbf{0.90}&  \textbf{0.70} 	&1.10&	2.47&	1.30\\
    & MC&0.98&	0.84&	 \textbf{0.96}& 	1.06&	0.96\\
    & MCR\_n&1.00&	0.82&	0.97&	0.95&	0.94\\
    & MCR\_{mp}&1.00&	0.93&	1.01&\textbf{0.80}&	 \textbf{0.94} \\ 
    & MCR\_{mpc} & 0.95 & 0.83 & 1.03 & 1.00 & 0.95 \\\hline
    \end{tabular}
    \end{center}
    \caption{Euclidean Errors in trajectory prediction over Crowd \& TownCentre datasets in meters. Observation length is 8 steps (3.2 seconds) and prediction length is 12 steps (4.2 seconds). }
    \label{tab:quant_table}
\end{table*}{}


Table \ref{tab:quant_table} summarizes our experimental evaluation over Zara and UCY datasets. 
Our models achieve state-of-the-art results in pedestrian trajectory prediction task. Taking the closest model, $MC$, it resembles the MX-LSTM design in terms of using mixture of pedestrian features, yet we are able to improve upon MX-LSTM \cite{hasan2018mx} by a margin of 8\% and 12\% in $MC$ and 10\% and 13\% in $MCR_{mp}$ over the average errors in ADE and FDE, respectively. 

In $MCR_{mp}$, we were able to achieve comparable results with SR-LSTM \cite{zhang2019sr}. They rely on sharing states between pedestrians using a message-passing mechanism that carries out the social information between pedestrian nodes in a global manner, meaning that they assume a fully-connected graph to refine the social interaction modeling through graphs. In Table \ref{tab:quant_table}, by comparing SR-LSTM to our message-passing model $MCR_{mp}$, we observe that both achieve approximate prediction errors, yet SR-LSTM induce higher message passing complexity due to using fully-connected graph and performing message passes pedestrian-wise to refine the social features representation. 
On the other hand, $MCR_{mp}$ encodes pedestrians awareness states into their social interaction using multiplication as previously discussed in Section \ref{sec:sri} without the need for iterative message-passing throughout the nodes. $MCR_{mp}$ model has better performance over the SR-LSTM method in Univ subset. The density of social interactions is the highest in Univ scenario with less impact from the static context on pedestrian paths. This is where the addition of social cues shows improvement in a highly interactive crowd, in comparison with the less interactive crowd as in Zara sets. The crowd in Zara is significantly influenced by the shop facade and therefore encoding the physical context features was more beneficial under these two subsets as we observe from $MCR_{mpc}$ performance. 

In general, the crowd in TownCentre induce a rather linear walking pattern with less socialization and less attention paid to the contextual remarks. Best results were obtained through $MCR_{mp}$, as it focuses on relational inference given the tracklets and vislets cues. This proves that our relational inference mechanism succeeded in inferring interactions even in the less interactive situation along with the inference performance in UCY-Univ where the crowd behaves like transient groups where the walks are interrupted by conversations.



\subsection{Ablation Studies}
We performed additional ablative experiments to quantitatively illustrate the importance of main grid components and attention mechanisms, by retraining the model each time without specific layer(s). We tested the impact of two components: the Grid-LSTM handling static grid shown earlier in Figure \ref{fig:gnn} and the soft-attention mechanism explained through Eq. \ref{eqn:soft_attn}. We followed the same k-fold cross-validation used for training the original MCR model on the same datasets: Zara and UCY.
Table \ref{tab:glstm_abl} clearly illustrates the improvement of our basic model, G-LSTM, upon the NoFrustum model \cite{hasan2018mx}. Our basic model only deploys one Grid-LSTM cell and relies on pedestrians positional trajectories. This design complexity resembles the NoFrustum model which is a simpler variant of MX-LSTM work. They only use pedestrians positions with LSTM cell. The usage of Grid-LSTM attains outperforming results over Zara and UCY in a row. 
We picked the NoFrustum variant as a baseline to compare with our basic model design G-LSTM, which excludes the GNN and SRI units from its pipeline and only take positional trajectories into single Grid-LSTM cell.

We tested neighborhood size on two values: 32 and 64 in the grid $G$. In our implementation, 64 local neighborhood creates a virtual neighborhood that covers a smaller area around pedestrian while 32 indicates that neighborhoods become coarser in size. Through experiments, we observed that the smaller the neighborhood, the higher the prediction error becomes. The errors increased by up to 15\% and 20\% in UCY-Univ over ADE and FDE respectively. This elevation has a special cause that is related to how our model benefits from encoding more pedestrian states together and understand that their cues are relevant to each other. However, in TownCentre, the pedestrians are far less communicative with each other, therefore, we noticed that ADE and FDE errors decreased by 10\% and 25\%, respectively. 

\begin{table}[t]
    \begin{center}
    \begin{tabular}{ccccccc}
    \hline
       Metric & Model & Zara1 & Zara2 & UCY-Univ & TownCentre & AVG\\ \hline
       ADE & NoFrustum \cite{hasan2018mx} & 0.63 & \textbf{0.36} & 0.51 & 1.70 & 0.50\\ 
         & SR-LSTM$_{\{ID:2\}}$ & 0.47 & 0.38 & 0.56 & 1.36 & 0.77\\
         & G-LSTM  &\textbf{0.46} & 0.38 &  \textbf{0.48} & \textbf{0.38} & \textbf{0.43}\\ \hline
        FDE & NoFrustum \cite{hasan2018mx} & 1.40& 0.84 &1.15 & 3.40& 1.13 \\ 
        &  SR-LSTM$_{\{ID:2\}}$ & 1.07  & 0.85 & 1.27& 2.47& 1.42  \\
        &  G-LSTM & \textbf{0.95}& \textbf{0.83} & \textbf{1.01}& \textbf{0.89} & \textbf{0.92} \\\hline
    \end{tabular}
    \end{center}{}
    \caption{Comparison of trajectory prediction errors between MX-LSTM NoFrustum model, SR-LSTM$_{\{ID:2\}}$ and our basic Grid-LSTM model, G-LSTM. Errors are displayed in meters. SR-LSTM$_{\{ID:2\}}$ is a model configuration which excludes social attention and selective message-passing routes between pedestrians.}
    \label{tab:glstm_abl}
\end{table}{}


\paragraph{Component Impact Analysis}
Before encoding the static context features in $MCR_{mpc}$, the virtual grid $G$ was considered an additional encoding layer that only transforms the multiple features together concerning the social interaction without adding a reference to the contextual interaction. Hence, it indicated low effectiveness and appeared as unnecessary to use the static grid to achieve noticeable improvements. Accounting for the semantic features in $G$ illustrates the reduction in FDE across all datasets and this proves the positive impact of this feature on predicting pedestrian trajectory at its final point being aware of their static surrounding settings. The addition of this component only increases the running time by 0.30 seconds compared to our socially-focused models.

\subsection{Qualitative Analysis}
\begin{figure}
    \subfloat[][Attention weights from $MCR_{mp}$ over the virtual grid in Zara1 scene]{\label{zara_htmap}
    \includegraphics[width=0.3\linewidth, height=0.3\linewidth]{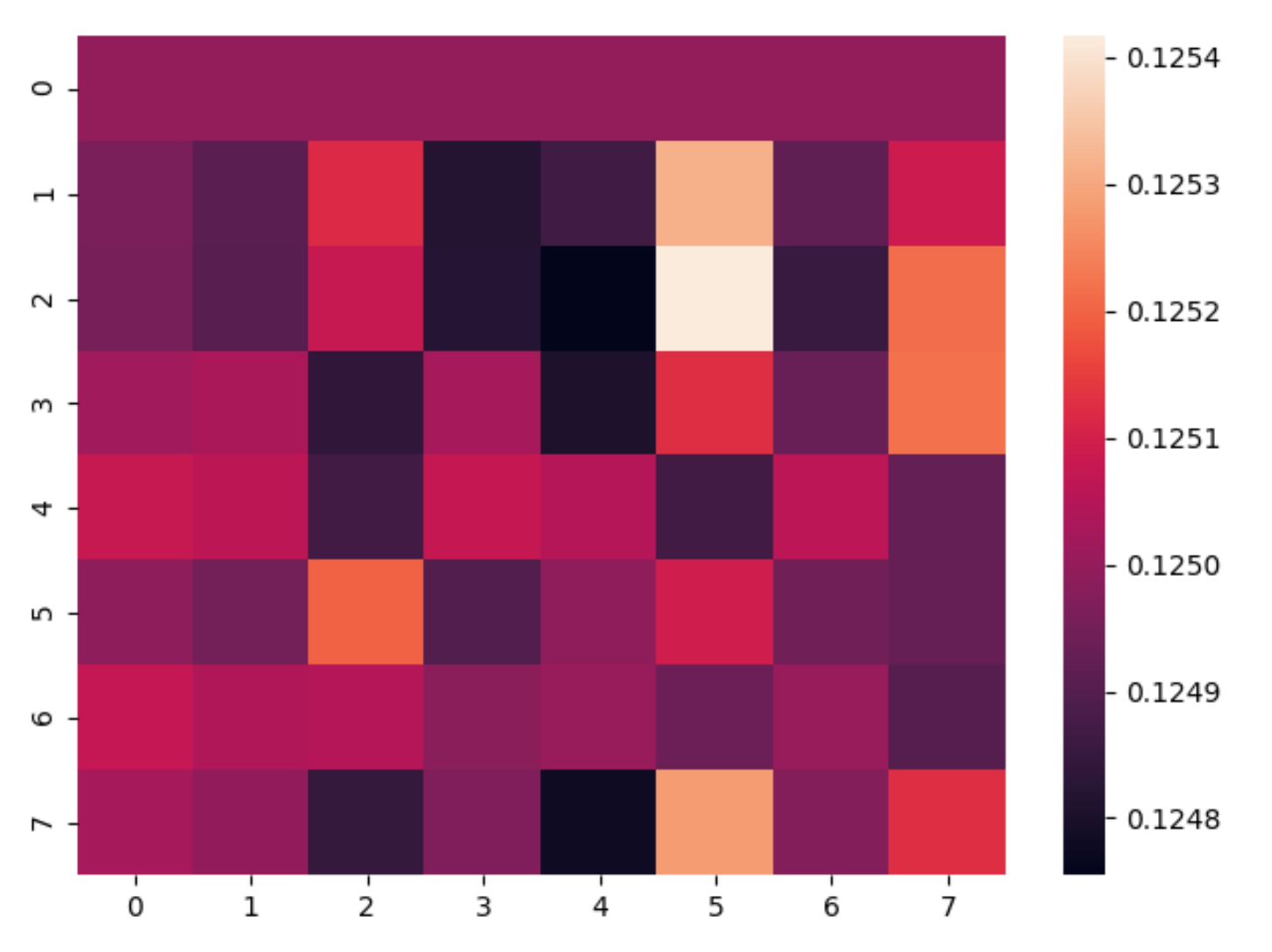}
    }
    \subfloat[][Maximum attention weight assigned to the neighborhood highlighted by the white patch, in Frame 8015 of Zara2 dataset.]{\label{zara_8015_maxadjmat}
    \includegraphics[width=0.3\linewidth, height=0.3\linewidth]{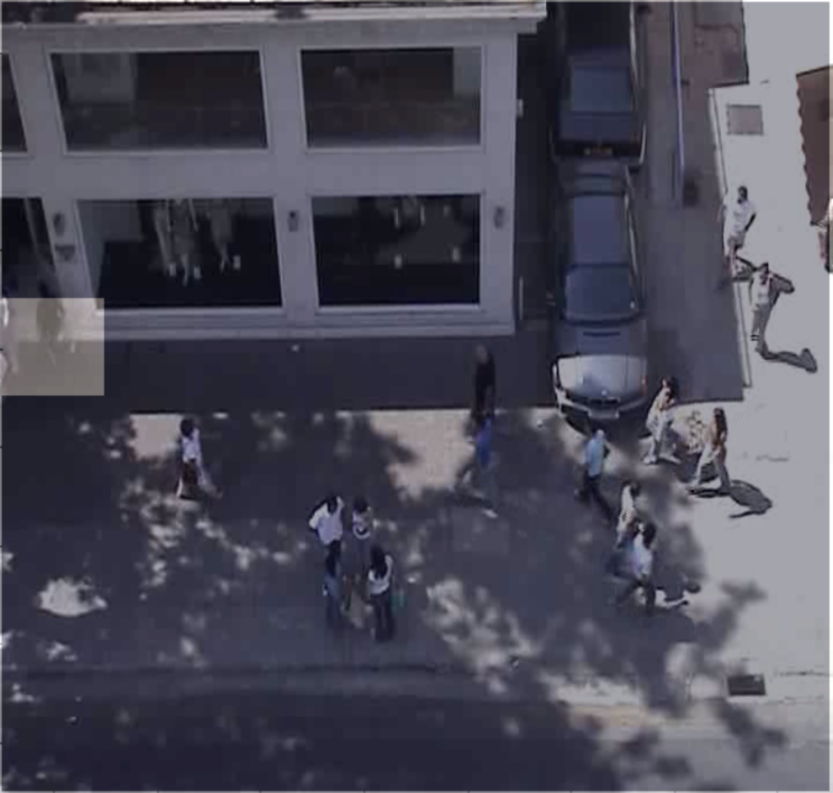}}
    \subfloat[][Failure case of assigning maximum importance to neighborhoods in the scene]{\label{zara_7815_maxadjmat}
    \includegraphics[width=0.3\linewidth, height=0.3\linewidth]{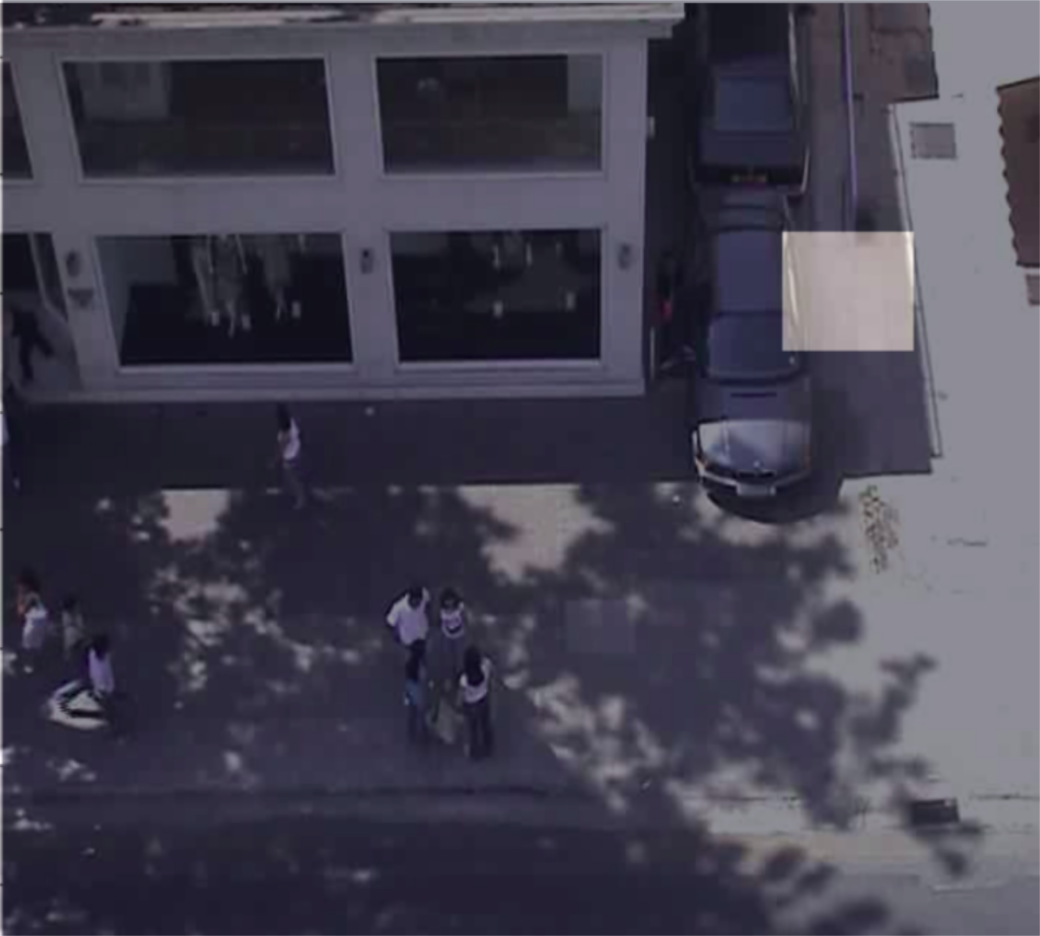}}
    \caption{Snapshots taken from our models to visualize the perception of neighborhoods relevant importance to pedestrians.}
    \label{fig:qualit_mcr}
\end{figure}

Figure \ref{zara_htmap} is a visualization of attention weights generated in $MCR_{mp}$. Attention weights, in this case, are distributed over the active part of the scene which includes the navigable space and social interaction between pedestrians. The dark squares include the least active areas where an obstacle is blocking motion. 


Moreover, the heatmap of adjacency matrix is a depiction of each neighborhood. These are assigned weighted values to show different impact on the crowd motion. This is manifested through the zones that highlight higher impact through the lighter shades of color, such as Zara shop entrance and the walkway at the right side. The lighter shade means that throughout the observed frames, these two regions were busy and have specific influence that causes change in pedestrians direction or dynamics.

Figure \ref{zara_7815_maxadjmat} presents failure case of estimating an active neighborhood under $MCR_n$. However, $MCR_{mpc}$ relates attention to physical space features. Figure \ref{zara_8015_maxadjmat}, illustrates that the maximum attention is given to the area through which pedestrian future path is passing. This illustration proves that our model is able to effectively recognize future neighborhoods for pedestrians as these neighborhoods presents target points that were regularly visited in the scene.


{}


\section{Conclusion}
In this work, we introduced a novel perspective on modeling social and contextual interactions as virtual neighborhoods with an estimation of its future relative importance values as a Gated Neighborhood Network (GNN). We developed the Social Relational Inference (SRI), a data-driven inference mechanism, to model the social interaction on graphs with no reliance on any metric assumption. Our approach outperformed state-of-the-art approaches that conduct pedestrian trajectory prediction with the help of several hand-engineered settings to model the crowd motion. 
In our future work, we plan to further improve the alignment between GNN and SRI components, so as to generate more realistic perception of the neighborhoods importance. Additionally, the current approach takes 3 to 4 seconds to run predictions over moderately crowded scene consisting of around 12 pedestrians. We will investigate methods to improve the training process in deep recurrent models such that it can eliminate network component complexity.



\bibliographystyle{splncs}
\bibliography{egbib}
\end{document}